\documentclass[preprint,review,12pt]{elsarticle}

\usepackage{color}

\usepackage{amssymb}
\usepackage{algorithm}
\usepackage{algorithmic}

\usepackage{amsmath}
\usepackage{multirow}
\usepackage{booktabs}
\usepackage{graphicx}
\usepackage{natbib}
\usepackage{tikz}
\usepackage{multirow}
\usepackage{bbding}

\usepackage{lineno}

\begin{document}

\begin{frontmatter}



\title{NeuroMoCo: A Neuromorphic Momentum Contrast Learning Method for Spiking Neural Networks}


\author[1,2]{Yuqi Ma}
\author[1,2]{Huamin Wang\corref{cor1}}
\cortext[cor1]{Corresponding author. Email: hmwang@swu.edu.cn (Huamin Wang)}
\author[1,2]{Hangchi Shen}
\author[1,2]{Xuemei Chen}
\author[1,2]{Shukai Duan}
\author[3]{Shiping Wen}

\affiliation[1]{
    organization={Southwest University},
    city={Chongqing},
    postcode={400715}, 
    country={China}
}

\affiliation[2]{
    organization={Chongqing Key Laboratory of Brain Inspired Computing and Intelligent Chips},
    city={Chongqing},
    postcode={400715}, 
    country={China}
}

\affiliation[3]{
    organization={University of Technology Sydney},
    addressline={Australian Institute of Artificial Ieintelligence}, 
    city={Sydney},
    postcode={2007}, 
    country={Australia}
}

\begin{abstract}
Recently, brain-inspired spiking neural networks (SNNs) have attracted great research attention owing to their inherent bio-interpretability, event-triggered properties and powerful perception of spatiotemporal information, which is beneficial to handling event-based neuromorphic datasets. 
In contrast to conventional static image datasets, event-based neuromorphic datasets present heightened complexity in feature extraction due to their distinctive time series and sparsity characteristics, which influences their classification accuracy.
To overcome this challenge, a novel approach termed \textbf{Neuromorphic Momentum Contrast Learning (NeuroMoCo)} for SNNs is introduced in this paper by extending the benefits of self-supervised pre-training to SNNs to effectively stimulate their potential. This is the first time that self-supervised learning (SSL) based on momentum contrastive learning is realized in SNNs. 
In addition, we devise a novel loss function named MixInfoNCE tailored to their temporal characteristics to further increase the classification accuracy of neuromorphic datasets, which is verified through rigorous ablation experiments.  Finally, experiments on DVS-CIFAR10, DVS128Gesture and N-Caltech101 have shown that NeuroMoCo of this paper establishes new state-of-the-art (SOTA) benchmarks: 83.6\% (Spikformer-2-256), 98.62\% (Spikformer-2-256), and 84.4\% (SEW-ResNet-18), respectively.
\end{abstract}




\end{frontmatter}


\section{Introduction}
\label{Introduction}
Spiking nueral networks (SNNs) have attracted a lot of research interest in recent years due to its bio-interpretable\cite{zhao2022spiking} and event-triggered properties. When executing SNNs on neuromorphic chips, the computation skips weight calculations corresponding to the spike "0" signal, requiring only accumulation of weights corresponding to the spike "1" signal\cite{pei2019towards,roy2019towards}. This significantly reduces power consumption compared to artificial neural networks (ANNs). Therefore, SNNs hold more potential to replicate the efficiency advantages of the human brain. In particular, the unique spatio-temporal perceptual properties of SNNs\cite{skatchkovsky2021spiking} give them intrinsic potential in complex pattern recognition tasks oriented to neuromorphic datasets. However, training SNNs for comparable performance on neuromorphic datasets remains challenging due to the gap between the expressive ability of 0/1 spiking signals and that of floating-point number signals in ANNs. To this end, some advanced works have considered and tried from different perspectives, including designing the residual connection structure of SNN\cite{fang2021deep,10428029,shen2024multi}, introducing the attention mechanism into SNN\cite{yao2021temporal,bernert2019attention}, and designing spiking neurons with richer neuronal dynamics\cite{fang2021incorporating,cheng2023meta}. Nevertheless, these efforts predominantly concentrate on architectural refinements and overlook potential solutions at the training method of SNNs.

Unsupervised representation learning has demonstrated notable advancements in ANNs, particularly in natural language processing (NLP), exemplified by GPT\cite{brown2020language} and BERT\cite{kenton2019bert}. This is because compared with supervised learning, unsupervised learning can use unlabeled large-scale data for training, learn the potential structure and pattern in the data, and thus improve the expression ability and generalization ability of the model. However, in computer vision (CV), supervised learning remains predominant due to the less discrete nature of visual signal space compared to language tasks. Self-supervised learning(SSL), a variant of unsupervised learning, leverages inherent data properties as a form of supervised signal for training. To effectively apply SSL to CV tasks, researchers have undertaken numerous investigations into image self-supervised representation learning, yielding remarkable outcomes. For instance, in some detection and segmentation tasks, MoCo\cite{he2020momentum} surpassed their supervised counterparts; SimCLR\cite{chen2020simple} has further narrowed the gap between unsupervised and supervised pre-training; and DINO\cite{caron2021emerging} introduced contrastive learning and knowledge distillation into SSL, resulting in significant performance enhancements.

In SNNs, except for SpikeGPT\cite{zhu2023spikegpt} in NLP tasks, and MAE\cite{he2022masked} used in Spikformer V2\cite{zhou2024spikformer}, the vast majority of works adopted supervised learning training methods, which means that there is a gap in SSL methods for CV tasks of SNN. Hence, it is a promising opportunity to leverage the advantages demonstrated by SSL in ANNs to enhance the potential of SNNs in addressing challenges associated with complex neuromorphic datasets (i.e.  datasets collected by event-based dynamic vision sensors). The development of DVS dynamic image (neuromorphic data collected by dynamic vision sensor cameras) classification not only provides new insights and technical support for the advancement of intelligent perception systems\cite{yang2023neuromorphic} but also holds extensive application prospects in fields such as autonomous driving\cite{chen2020event} and drone navigation\cite{salvatore2020neuro}.

In this paper, based on MoCo paradigm\cite{he2020momentum,chen2020improved,chen2021empirical}, a SNN-oriented Neuromorphic momentum contrast learning method (NeuroMoCo) is proposed to enhance the accuracy of DVS dynamic image classification, which can be used as a self-supervised pre-training framework for spiking convolution and spiking Transformer structures. Here, data augmentation techniques in NDA\cite{li2022neuromorphic} are improved to effectively enhance the diversity of positive and negative samples. At the same time, a new loss function named MixInfoNCE is designed by timing characteristics to increase the classification accuracy of neuromorphic datasets. In the end, to validate the NeuroMoCo of this paper, rich experiments are conducted on mainstream neuromorphic datasets DVS-CIFAR10, DVS128Gesture, and N-Caltech101. The experimental results show that:1) For DVS-CIFAR10, DVS128Gesture, and N-Caltech101, integrating with our NeuroMoCo framework of this paper, SEW-ResNet-18 and Spikformer-2-256 can 
respectively attain 81.50\%, 97.92\%, 84.35\% and 83.60\%, 98.62\%, 81.62\% classifition accuracy, surpassing those achieved through random initialization training;
2) The ablation experiments of the loss function demonstrate the viability and efficacy of MixInfoNCE loss function;
3) Compared with current leading methodologies, our NeuroMoCo approach establishs state-of-the-art (SOTA) benchmarks on DVS-CIFAR10, DVS128Gesture, and N-Caltech101. The contributions can be summarized as follows:
\begin{itemize}
\item We construct a dynamic dictionary, which integrates an automatically updating queue mechanism and an encoder based on momentum sliding average optimization. Based on this, we propose a SNN-oriented neuromorphic momentum contrast learning method (NeuroMoCo) to pretrain SNN model.
\item We present a pre-processing method for neuromorphic datasets, and improve the data augmentation technique in NDA\cite{li2022neuromorphic} to effectively enhance the diversity of positive and negative samples.
\item According to the timing characteristics of neuromorphic datasets, we design a new loss function named MixInfoNCE, which is verified through rigorous ablation experiments.
\item We conduct experiments on DVS-CIFAR10, DVS128Gesture, and N-Caltech101 to test the advances
of NeuroMoCo. It is worth mentioning that compared with current leading methodologies, our NeuroMoCo approach establishs state-of-the-art (SOTA) benchmarks on each datasets, denoted as 83.6\% (Spikformer-2-256), 98.62\% (Spikformer-2-256) and 84.4\% (SEW-ResNet-18), respectively.
\end{itemize}

The remainder of this article is structured as follows: Section 2 provides a concise overview of relevant prior work.   In Section 3, we present the proposed method with details.  Section 4 delineates the experimental methodologies employed to assess the efficacy of our proposed method and the associated loss function, and presents the ensuing experimental results.   Finally, in Section 5, we draw conclusions grounded in our research findings.
\section{Related Work}\label{sec:Related work} 
\subsection{Spiking Neural Networks (SNNs)}
SNNs are widely regarded as the third generation of neural networks, succeeding the McCulloch-Pitt perceptron and ANNs, primarily owing to their inherent biological plausibility\cite{taherkhani2020review}. Unlike traditional ANNs, which rely on continuous floating-point numbers to encode information, neurons in SNNs (LIF neurons\cite{teeter2018generalized}, PLIF neurons\cite{fang2021incorporating} etc.) encode continuous input signals into binary spike sequences (0/1) through neuronal firing, which makes SNNs use discrete and sparse spike sequences to characterize information.     Furthermore, SNNs introduce a temporal dimension, according with the special space-time characteristics of biological organism. The prevailing direct training\cite{wu2019direct} is adopted to construct SNNs in this study, as our task necessitates direct pre-training and fine-tuning of the SNN model.   While numerous effective models and techniques have been successfully used in SNNs\cite{su2023deep,liao2024spiking,kim2022beyond}, the domain of SSL within SNNs remains relatively underexplored.   In this investigation, our objective is to investigate the viability and efficacy of transferring self-supervised pre-training methodologies to CV tasks in SNNs.    Such an exploration seeks to enhance the representational capabilities of SNNs on intricate neuromorphic datasets.
\subsection{Self-Supervised Learning (SSL)}
SSL has demonstrated its effectiveness in CV and NLP tasks in ANNs. This learning paradigm encompasses two primary branches: masked autoencoder\cite{he2022masked} and contrastive learning\cite{le2020contrastive}. The former reconstructs masked input to learn feature representations, while the latter enhances generalization and performance by comparing similarities and differences among sample pairs. Recent advancements include MoCo series\cite{he2020momentum,chen2020improved,chen2021empirical}, which introduce contrastive learning to computer vision using a momentum update strategy, and DINO\cite{caron2021emerging}, which leverages a Transformer architecture and self-supervised self-representation learning to maximize mutual information in image feature representations, enriching feature representation. Here, we propose to explore self-supervised momentum contrastive learning in SNNs for the first time.
\section{Method}
Towards neuromorphic datasets, a self-supervised pre-training framework named NeuroMoCo is specifically designed for SNNs, capable of accommodating both spiking convolution and spiking Transformer architectures, thereby enhancing the expressive ability and generalization ability of SNN models. The details of NeuroMoCo are despited as follows.
\subsection{Overall Architecture}
\begin{figure}
    \centering
    \includegraphics[scale=0.35]{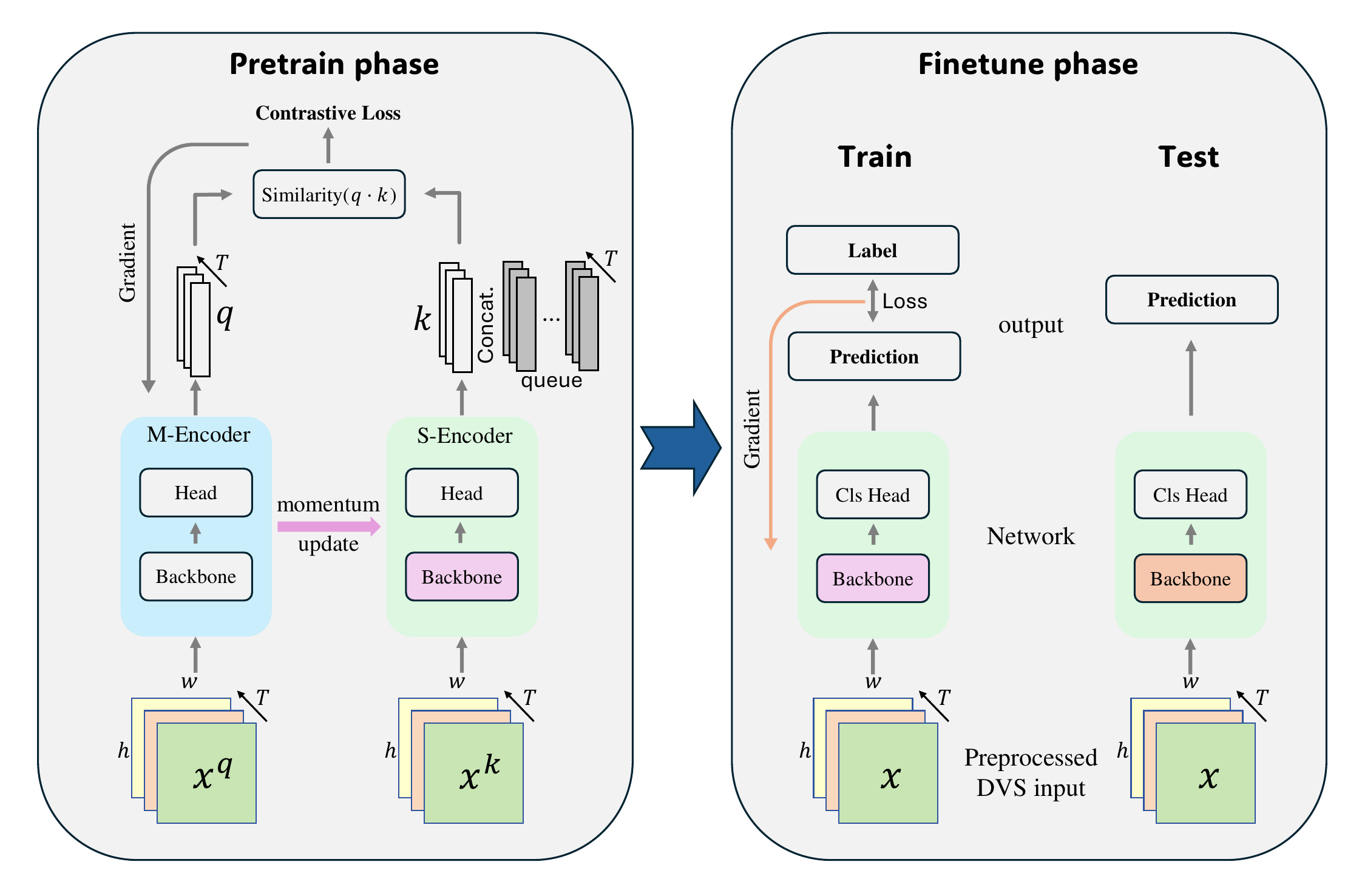}
    \caption{Overview of NeuroMoCo and subsequent fine-tune. The NeuroMoCo includes an automatically updating queue and the S-Encoder based on momentum sliding average optimization. $x_q$ and $x_k$ are obtained by processing one data sampled from the neuromorphic dataset using two randomly different data augmentation methods. T represents the time dimension of DVS data.}
    \label{fig:Figure1}
\end{figure}

Regarding NeuroMoCo, we will present the details following the "from pre-training to fine-tuning" paradigm. In the pre-training phase (figure \ref{fig:Figure1}, left), the input $x_q$ of the master encoder (M-Encoder) and the input $x_k$ of the subordinative encoder (S-Encoder) are obtained by processing one data sampled from the neuromorphic dataset using two randomly different data augmentation methods. It is important to note that the neuromorphic dataset inherently possesses a temporal dimension, which we represent as T. The neuromorphic dataset during sampling is obtained after preprocessing the original neuromorphic data, and the specific preprocessing method is described in 3.2.

Given the input $x_q$, M-Encoder encodes it to contextualized vector representations q, while the S-Encoder, with the same structure and initialization as the M-Encoder, encodes the input $x_k$ into a contextualized vector representation k. k is homologous to q and thus serves as a positive sample of q. Concurrently, contrastive learning necessitates assessing the similarity between positive and negative samples, and acquiring good features relies on abundant negative samples.    To this end, we constructed a pool of negative samples for q utilizing a dynamic queue, denoted as \{k0, k1, k2,...\}. First of all, we initialize an empty queue of fixed size.      At the beginning of training, the queue lacks sufficient negative samples for comparison with the positive samples, hence, we commence with random negative pairs.

Subsequently, after processing each sample, the feature representation of that sample is appended to the queue, while the earliest feature representation is evicted from the queue, thereby maintaining a constant queue size.  Such a dynamic update method is based on the fact that the earliest feature representation in the queue is the most outdated and the most inconsistent with the feature representation obtained by the latest encoding.      Consequently, as training proceeds, the feature representations in the queue undergo continual updates, thus constituting a feature pool comprising a specific number of negative samples. Certainly, in implementation, the process operates in batches, implying that a batch of size samples is processed at a time.   Subsequent to acquiring the negative sample pool, the positive sample k and the dynamic queue of negative samples are concatenated to form the positive and negative sample space of q, denoted as K=\{k, k0, k1, k2,...\}.  It is important to emphasize that q and the samples within K remain temporal samples, with T still representing their time dimension.

On this basis, the contrastive loss function is used to represent the difference between q and its positive and negative samples, and the value of function is small when q is similar to its positive sample but not similar to the remaining negative samples (see 3.4 for the contrastive loss function).      Then, the contrastive loss was gradually reduced based on gradient descent and back propagation algorithm, and the parameters of the M-Encoder were updated;     for S-Encoder, we adopted a momentum update strategy because the negative samples processed by S-Encoder came from several previous small batches.       The strategy can be formulated as follows:
\begin{align*}
    \mathbf{\theta_k}&=m\mathbf{\theta_k}+(1-m)\mathbf{\theta_q}\tag{1}
    \label{eq:momentum update}
\end{align*}
where $\mathbf{\theta_q}$ and $\mathbf{\theta_k}$ denote the parameters of the M-Encoder and S-Encoder, respectively, and m$\in$[0,1) signifies the momentum update coefficient.     The utilization of momentum update enables the S-Encoder to maintain the direction consistency of parameter updating, thereby leveraging historical information more effectively and acquiring a more discriminative feature representation.

After pre-training with our NeuroMoCo method, the S-Encoder replaces the Head network behind Backbone network with a classification Head network in the fine-tuning stage (figure \ref{fig:Figure1}, right), so as to conduct subsequent training and testing of classification tasks on specific neuromorphic datasets.    It should be noted that the treatment of T when calculating the loss in this stage follows the MAC paradigm consistent with the contrastive loss function, as detailed in 3.4.
\subsection{Neuromorphic Data Preprocess}
\begin{figure}
    \centering
    \includegraphics[scale=0.8, width=1.0\textwidth]{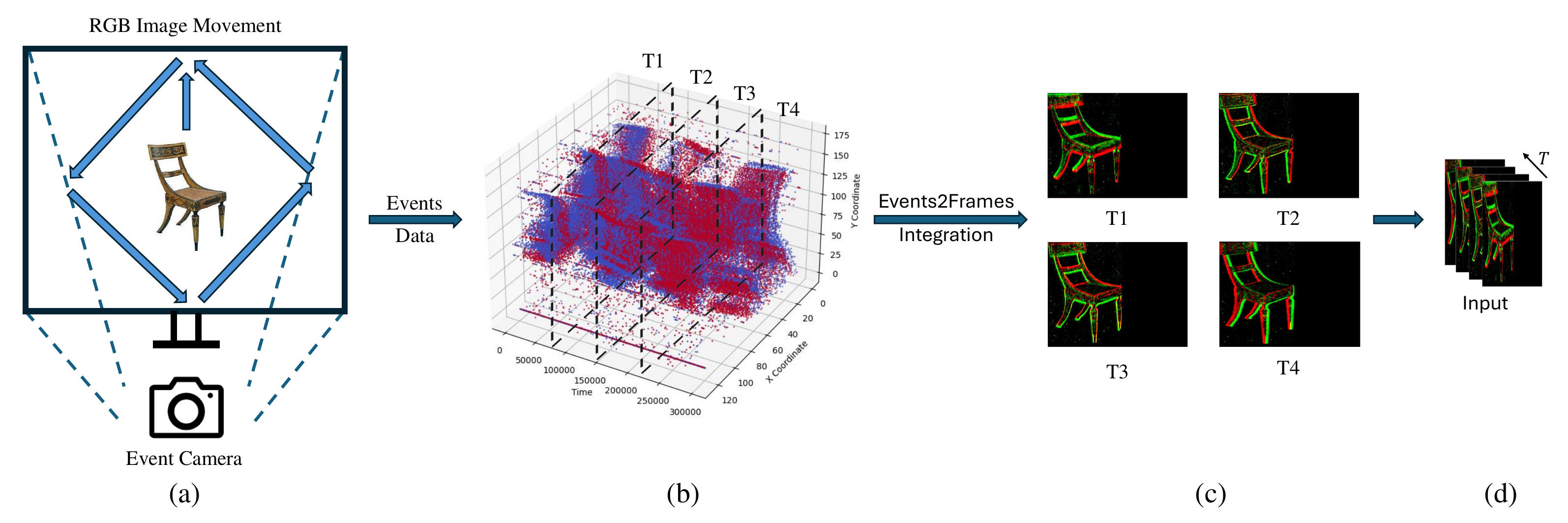}
    \caption{Collection and preprocessing of neuromorphic data. The DVS camera collects sparse event data, which are integrated into time frames and stored in a large multi-dimensional tensor according to time series.}
    \label{fig:Figure2}
\end{figure}
In this paper, the neuromorphic datasets we use are all event-based datasets collected by Dynamic Vision Sensor (DVS) cameras, which we Uniformly name as DVS data.        Taking a chair image in the N-Caltech101 dataset as an example, the collection process of DVS data is shown in Figure \ref{fig:Figure2} (a), which makes the static RGB image move along a certain trajectory, and the DVS camera captures this process and outputs the event stream data.      When employing a DVS camera for data collection, each pixel operates independently and asynchronously.      When the pixel is changed, it will show a positive active state, otherwise it will keep a negative silent state.        Consequently, the resulting event stream data (as shown in Figure \ref{fig:Figure2} (b)) manifests as time sequential and sparse.   Due to it is too sparse, direct feature extraction becomes exceedingly challenging, necessitating preprocessing of neuromorphic data.

As mentioned above, the event stream data output by DVS camera is time-series.    Therefore, we first divide the sparse event stream data into time Windows according to the time sequence (in Figure \ref{fig:Figure2} (b), it is divided into four time windows T1, T2, T3, T4).    Then the events are grouped according to the time window.    For each time window, we will find all the events that fall within the time window and group them into a list.    For the list of events in each time window, we need to integrate these events into the corresponding time frame (Figure \ref{fig:Figure2} (c)),which is implemented by counting the number, spatial distribution and polarity of events in the time window.    Different polar events are stored in different channels in the time frame.  Finally, we stored the integrated time frames in each time window in a large multi-dimensional tensor according to the time sequence to obtain the preprocessed neuromorphic data (Figure \ref{fig:Figure2} (d)).
\subsection{Backbone Network}
To bolster the credibility of our final conclusion, we opt for widely recognized convolutional architecture and Transformer architecture as backbones when employing provided NeuroMoCo for pre-training.
\begin{figure}
    \centering
    \includegraphics[scale=0.3]{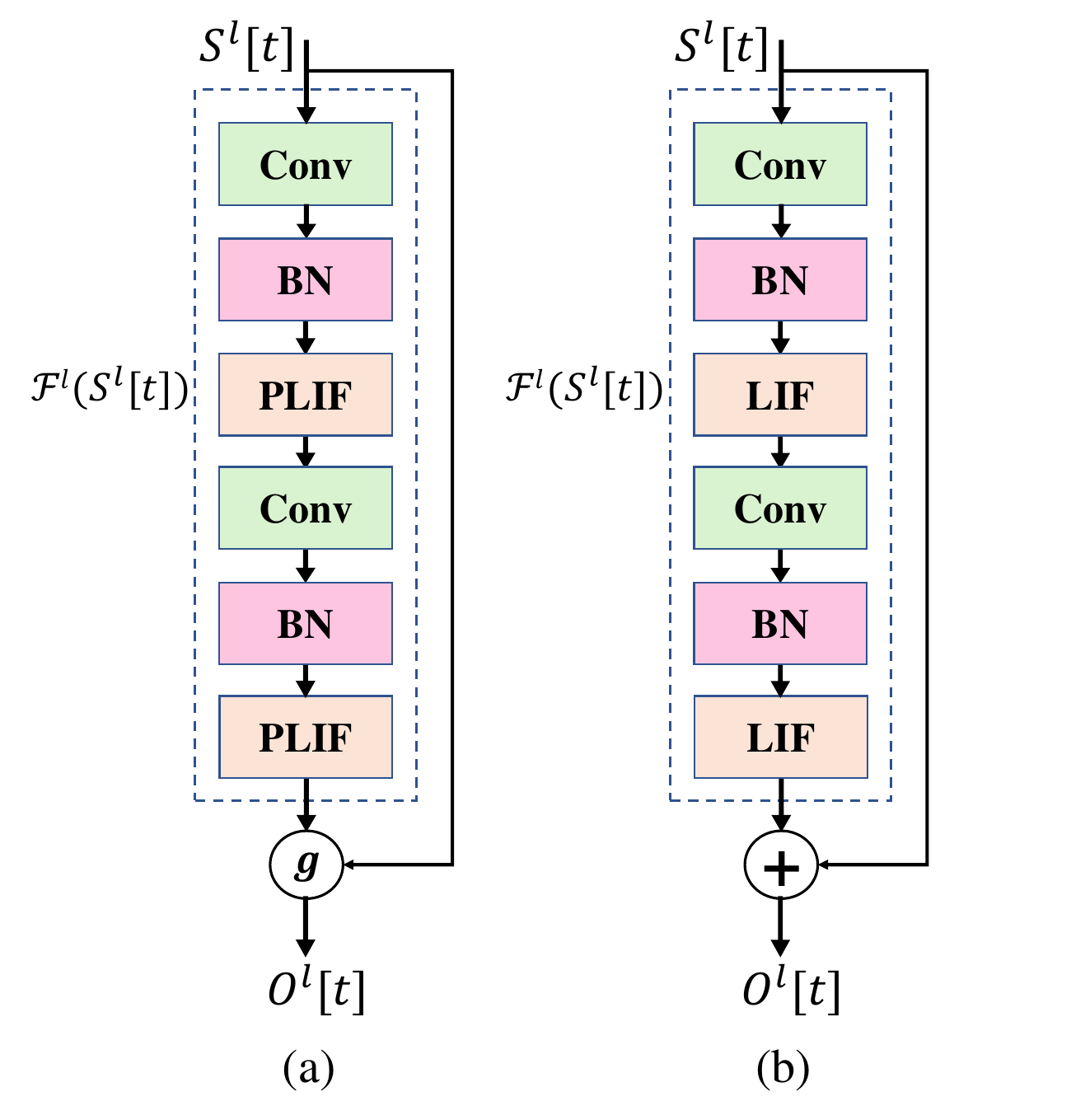}
    \caption{Overview of Spike-Element-Wise block. We use ADD as the element-wise function (g) and substitute the original PLIF neurons with more versatile LIF neurons.}
    \label{fig:Figure3}
\end{figure}

Specifically, for the convolutional architecture, following SEW-ResNet\cite{fang2021deep}, we constructed the SEW-RESNET-18 model as the Backbone using Spike-Element-Wise blocks (Figure \ref{fig:Figure3} (a)).   Notably, as depicted in Figure \ref{fig:Figure3} (b), within the Spike-Element-Wise blocks, we directly employ ADD as the primary element-wise function (g). Additionally, to mitigate the specific effects of neurons and promote universality,   we substituted the original PLIF neurons with more versatile LIF neurons.     Consequently, we establish the Backbone of the convolutional architecture on this foundation.

For the Backbone of Transformer architecture, we construct the Spikformer-2-256 model, building upon Spikformer\cite{zhou2022spikformer}.      This means that two spikformer encoder blocks are included and the feature embedding dimensions are 256.      Furthermore, the patch size is set at 16 × 16, and the number of heads in the Spiking Self-Attention (SSA) module is uniformly configured to 16.      It is imperative to underscore that, given the nature of neuromorphic data, the input comprises two channels representing the positive and negative polarities of the data.
\subsection{Contrastive Loss}
The contrastive loss function measures the similarity of pairs of samples in a representation space, and its value needs to be smaller when the positive pairs exhibit higher similarity while the negative pairs display lower similarity.     InfoNCE\cite{oord2018representation} is a more mainstream form of contrastive loss function.     It obtains the similarity matrix through dot product operation, so as to measure the similarity between sample pairs.     Based on InfoNCE and the unique time series characteristics of neuromorphic data, a contrastive loss function named MixInfoNCE is designed in this paper.
\begin{figure}
    \centering
    \includegraphics[width=1.0\textwidth]{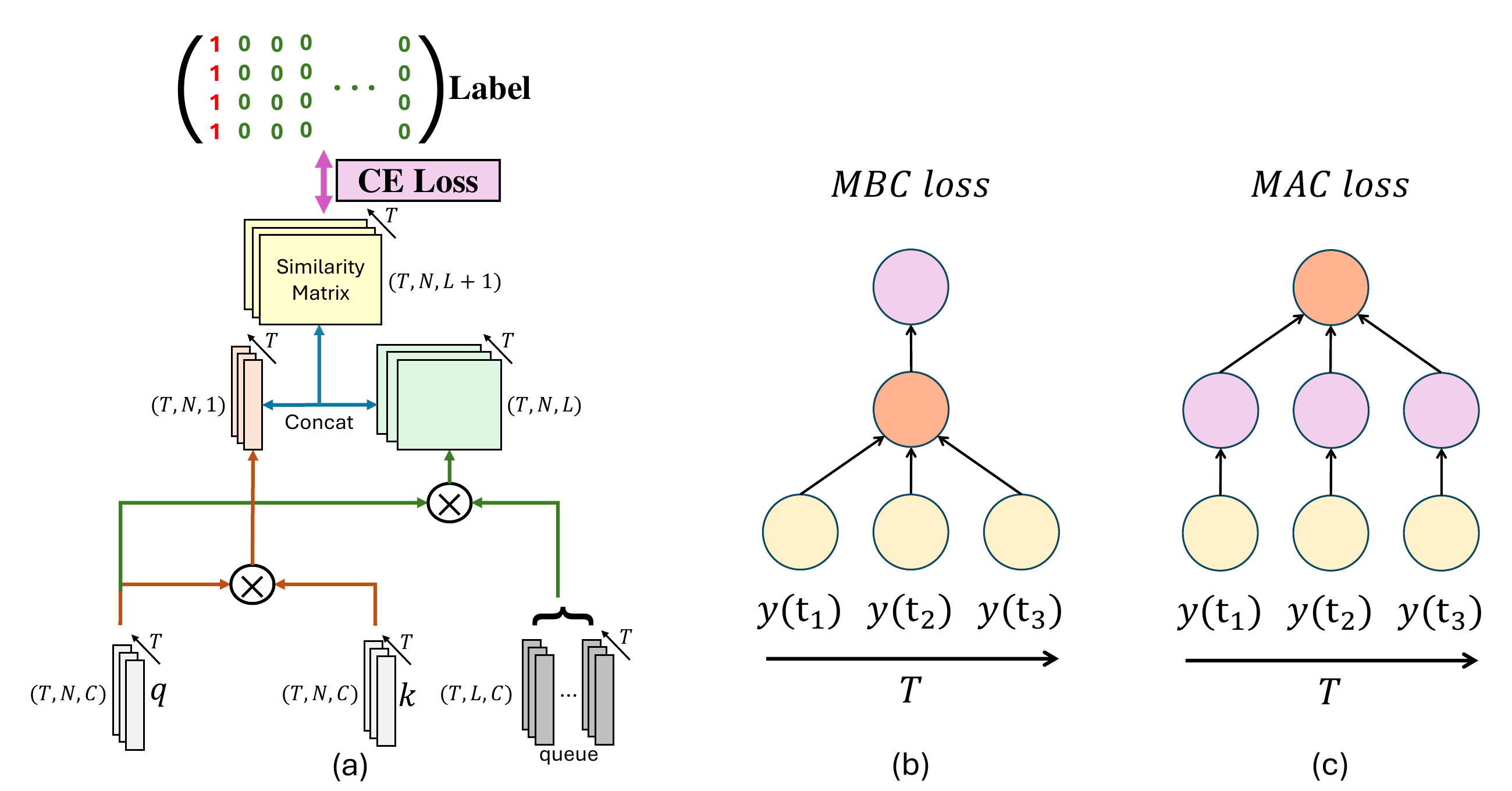}
    \caption{The principle diagram of MixInfoNCE. Following InfoNCE, we obtain the similarity matrix with time dimension T. For T, our MixInfoNCE adopts the strategy of MBC\&MAC mixed paradigm.}
    \label{fig:Figure4}
\end{figure}

As shown in Figure \ref{fig:Figure4} (a), q(T,N,C) and k(T,N,C) are the vector representations output by M-Encoder and S-Encoder, respectively.    k is the positive sample of q. queue(T,L,C) is a negative samples queue of q composed of k stored in history, where T represents the unique time dimension of neuromorphic data, N represents the number of samples processed in batches, C represents the number of vector-encoded channels, and L represents the fixed length of the queue. The overarching objective is to render positive sample pairs entirely similar and negative sample pairs entirely dissimilar.  In constructing the labels, positions corresponding to the positive similarity matrix are set to 1, while positions corresponding to the negative similarity matrix are set to 0.  Subsequently, the cross-entropy Loss (CE Loss) between the similarity matrix and the label is computed.  This process can be formalized as follows: 
\begin{align*}
    \mathcal{L}_\text{q-InfoNCE}&=l_{CE}(SimilarityMatrix,y_{gt})\\
    &=-\log\frac{\exp(q\cdot k/\tau)}{\exp(q\cdot k/\tau)+\sum_{i=0}^{K-1}\exp(q\cdot k_i/\tau)}\tag{2}
    \label{eq:InfoNCE}
\end{align*}
where $\tau$ denotes the temperature coefficient hyperparameter and $y_{gt}$ represents the true label.

However, the conventional InfoNCE does not inherently incorporate the time dimension in its computation.      Thus, it becomes necessary to handle the time dimension specific to neuromorphic data separately.      In SNN, the loss function typically follows the "mean before criterion" (MBC) paradigm, as illustrated in Figure \ref{fig:Figure4} (b), that is, the mean operation is used to eliminate the time dimension before calculating the difference between the prediction result and the label.  It can be expressed as follows:
\begin{align*}
    \mathrm{\textit{loss}}_{MBC}=l_{CE}(mean(f_{SNN}(x(t))),y_{gt})\tag{3}
    \label{eq:MBC}
\end{align*}
where $x(t)$ represents the time frame input of SNN, $f_{SNN}(\textbf{·})$ signifies the SNN encoding operation, $mean(\textbf{·})$ denotes the averaging across the time dimension, and $l_{CE}(\textbf{·})$ represents the computation of cross-entropy loss. Moreover, drawing inspiration from \cite{deng2022temporal}, we introduce a novel paradigm for the loss function termed "mean after criterion" (MAC), as depicted in Figure \ref{fig:Figure4} (c). In this paradigm, the time dimension is integrated into the calculation of the disparity between the predicted result and the label. It can be expressed as follows:
\begin{align*}
    \mathrm{\textit{loss}}_{MAC}=mean(l_{CE}(f_{SNN}(x(t)),y_{gt})).\tag{4}
    \label{eq:MAC}
\end{align*}

Finally, we aim to achieve the interaction of local and global information  from the time dimension for superior performance. To this end, we devised the MixInfoNCE, formulated as follows:
\begin{align*}
    \mathcal{L}_\text{MixInfoNCE}&=\alpha \textit{loss}_{MBC}+\beta \textit{loss}_{MAC}\\
    &=\alpha l_{CE}(mean(SimilarityMatrix,y_{gt})\\
    &+\beta mean(l_{CE}(SimilarityMatrix,y_{gt}))\tag{5}
    \label{eq:MixInfoNCE}
\end{align*}
where $\alpha$ and $\beta$ are hyperparameters, with $\alpha+\beta=1$. In the experimental section, we assess its effectiveness through ablation experiments.
\section{Experiments and Analysis}
Experiments and analysis are given in this section to visually verify the above methodology, which is divided into two parts: experimental details and experimental results.
\subsection{Experimental Details}
In this part, we will elaborate on the experimental methods and details, focusing on aspects such as datasets, experimental settings, and more.
\subsubsection{Datasets}
This paper targets event-based neuromorphic datasets, hence we select three commonly utilized DVS datasets: DVS-CIFAR10, DVS128Gesture, and N-Caltech101.  

\textbf{DVS-CIFAR10} is a neuromorphic dataset based on CIFAR10, wherein pulse events of image samples are captured utilizing a DVS camera.      This dataset comprises 9,000 training samples and 1,000 testing samples.

\textbf{DVS128Gesture} is a gesture recognition dataset collected by a DVS128 dynamic vision sensor, covering 11 distinct gesture categories performed by 29 participants across 3 varying lighting conditions.

Akin to DVS-CIFAR10, the \textbf{N-Caltech101} dataset is an extension of the Caltech101 static image dataset.      Caltech101, renowned as a classic image classification dataset, encompasses image samples across 101 object categories, with each category containing approximately 50 to 800 image samples.

These datasets are captured by dynamic vision sensor cameras and exhibit characteristics such as time sequence and high sparsity.
\subsubsection{DVS Data Augmentation}
During pre-training with our NeuroMoCo method, data augmentation on the DVS data is necessary to construct positive-negative sample pairs.   It has been demonstrated in NDA that pixel value-based augmentation is not suitable for DVS data.   Consequently, our DVS data augmentation method relies on geometric-based augmentation.

In general, building upon the DVS data augmentation method employed by Spikformer\cite{zhou2022spikformer}, we make specific enhancements to suit our requirements.    Particularly, considering that the resolution of the data in N-Caltech101 differs from other datasets, and ensuring consistency in resolution across all data is essential for comparing positive and negative samples pairs.    Therefore, we employ Resize to augment the data of N-Caltech101.    Additionally, we introduce vertical shear transformation (ShearY) and random horizontal flip to further enrich the data augmentation strategy, ensuring its comprehensiveness.
\begin{table}[htbp]
\centering
\caption{Parameters of Pre-Training and Fine-Tuning Phase}
\begin{tikzpicture}
\draw (0,0) grid[xstep=2.2cm,ystep=1cm] (13.2,3);
\draw (0,3) -- (2.2,2);
\node at (1,2.5) {};
\node at (3.3,2.5) {time step};
\node at (5.5,2.25) {update(m)};
\node at (5.5,2.75) {momentum};
\node at (7.7,2.5) {batch size};
\node at (9.9,2.5) {epoch};
\node at (12.1,2.75) {learning};
\node at (12.1,2.25) {rate};

\node at (3.3,1.5) {16};
\node at (5.5,1.5) {0.999};
\node at (7.7,1.5) {32};
\node at (9.9,1.5) {200};
\node at (12.1,1.5) {0.03};

\node at (3.3,0.5) {16};
\node at (5.5,0.5) {-};
\node at (7.7,0.5) {16};
\node at (9.9,0.5) {100};
\node at (12.1,0.5) {0.001};

\node at (1.1,1.5) {Pre-Train};
\node at (1.1,0.5) {Fine-Tune};
\node at (0.6,2.3) {Phase};
\node at (1.4,2.7) {Setup};
\end{tikzpicture}
\label{tab:Parameters}
\end{table}
\subsubsection{Pre-Training and Fine-Tuning Setup}
We utilize NeuroMoCo to pretrain the Backbone networks on the synthetic neuromorphic dataset, which is composed of CIFAR10, N-Caltech101, and DVS128Gesture.      It is important to note that our self-supervised pre-training does not rely on labels.      Subsequently, we append a classification head to the Backbone and conduct supervised fine-tuning training and testing on CIFAR10, N-Caltech101, and DVS128Gesture, respectively.     During pre-training, owing to task requirements, we only retain two random views obtained from data augmentation, while in the fine-tuning phase, all data augmented views are preserved.     Throughout our experiments, we maintain a uniform resolution of 128×128.     The hyperparameter settings during the experiment are detailed in Table 1.

For pre-training, we choose Stochastic Gradient Descent (SGD) with weight decay of 1e-4 and momentum of 0.9 for optimization, and adopt MultiStepLR strategy for learning rate scheduling.  In the fine-tuning stage, we use AdamW with a weight decay of 0.06 for optimization, and the learning rate first warms up for 30 epochs and then decays following the CosineAnnealingLR strategy.
\subsection{Experimental Results}
In this subsection, we commence by conducting ablation experiments to probe the effectiveness of the designed loss function and the proposed NeuroMoCo.    Subsequently, the model performance is evaluated on DVS-CIFAR10, DVS128Gesture as well as N-Caltech101, and compared with some SOTA works in related fields to highlight the effect of our NeuroMoCo method.
\subsubsection{Ablation Experiment}
The key improvement of MixInfoNCE lies in the modification of the paradigm of the original InfoNCE when computing the cross-entropy loss.       To ascertain its effectiveness and advantages, we carry out extensive ablation experiments on the loss function across three neuromorphic datasets using models of two architectures: spike convolution and spike transformer.     To maintain consistency with subsequent experiments, during the ablation experiment, we directly use the models SEW-ResNet-18 and Spikformer-2-256 employed in the fine-tuning stage , and the time step is also uniformly set to 16.
Firstly, we use the CE loss of MBC paradigm as the loss function ($\textit{loss}_{MBC}$), and conduct direct training and testing on DVS-CIFAR10, DVS128Gesture and N-Caltech101, respectively, so as to obtain a set of benchmarks for ablation experiments. Then, under the same experimental setup, we replace the loss function with the CE loss of MBC and MAC mixed paradigm ($\alpha \textit{loss}_{MBC}+\beta \textit{loss}_{MAC}$, this paper only considers the case of $\alpha=\beta$). The same experiments were performed again on each corresponding dataset.
\begin{table}
    \centering
    \caption{Comparison of MBC loss with MBC\&MAC mixed loss}
    \begin{tabular}{cccc}
        \toprule
        Models & Datasets & Mixed Loss & Acc \\
        \hline
        \multirow{6}{*}{SEW-ResNet-18} & \multirow{2}{*}{DVS-CIFAR10} & \XSolidBrush & 78.00 \\
         &  & \Checkmark & 81.00 \\
        \cmidrule(lr){2-4} 
         & \multirow{2}{*}{DVS128Gesture} & \XSolidBrush & 95.83 \\
         &  & \Checkmark & 96.18 \\
         \cmidrule(lr){2-4}
          & \multirow{2}{*}{N-Caltech101} & \XSolidBrush & 80.52 \\
         &  & \Checkmark & 80.74 \\
        \hline
        \multirow{6}{*}{Spikformer-2-256} & \multirow{2}{*}{DVS-CIFAR10} & \XSolidBrush & 80.90 \\
         &  & \Checkmark & 81.60 \\
        \cmidrule(lr){2-4}
         & \multirow{2}{*}{DVS128Gesture} & \XSolidBrush & 96.87 \\
         &  & \Checkmark & 97.57 \\
         \cmidrule(lr){2-4}
          & \multirow{2}{*}{N-Caltech101} & \XSolidBrush & 79.86 \\
         &  & \Checkmark & 80.53 \\
        \bottomrule
    \end{tabular}
    \label{tab:loss}
\end{table}

\begin{table}
    \centering
    \caption{Ablation study results on NeuroMoCo.* denotes self-implementation results by \cite{zhou2022spikformer}.}
    \begin{tabular}{cccc}
        \toprule
        Models & Datasets & NeuroMoCo & Acc \\
        \hline
        \multirow{6}{*}{SEW-ResNet-18} & \multirow{2}{*}{DVS-CIFAR10} & \XSolidBrush & 78.00 \\
         &  & \Checkmark & 81.50 \\
        \cmidrule(lr){2-4} 
         & \multirow{2}{*}{DVS128Gesture} & \XSolidBrush & 95.83 \\
         &  & \Checkmark & 97.92 \\
         \cmidrule(lr){2-4}
          & \multirow{2}{*}{N-Caltech101} & \XSolidBrush & 80.52 \\
         &  & \Checkmark & 84.35 \\
        \hline
        \multirow{6}{*}{Spikformer-2-256} & \multirow{2}{*}{DVS-CIFAR10} & \XSolidBrush & 80.90 \\
         &  & \Checkmark & 83.60 \\
        \cmidrule(lr){2-4}
         & \multirow{2}{*}{DVS128Gesture} & \XSolidBrush & $96.87^*$ \\
         &  & \Checkmark & 98.62 \\
         \cmidrule(lr){2-4}
          & \multirow{2}{*}{N-Caltech101} & \XSolidBrush & 79.86 \\
         &  & \Checkmark & 81.62 \\
        \bottomrule
    \end{tabular}
    \label{tab:neuromoco}
\end{table}

The experimental results shown in Table 2 indicate that when CE loss of MBC and MAC mixed paradigm is used, the performances of SEW-ResNet-18 and Spikformer-2-256 on DVS-CIFAR10, DVS128Gesture, and N-Caltech101 surpass those achieved with the CE loss of MBC paradigm alone.

Likewise, in the ablation experiments regarding NeuroMoCo, as a set of baselines, We first perform a group of experiments using SEW-ResNet-18 and Spikformer-2-256 without NeuroMoCo across all three datasets. Then, we compare the effects of SEW-ResNet-18 and Spikformer-2-256 with and without the NeuroMoCo method on the three neuromorphic datasets, respectively.  The experimental results are summarized in Table 3.
\begin{table}
    \centering
    \caption{Comparison of performance between our NeuroMoCo and current state-of-the-art (SOTA) methods on neuromorphic datasets.}
    \begin{tabular}{cccccccc}
        \toprule
        \multirow{2}{*}{Methods} & \multirow{2}{*}{Spikes} & \multicolumn{2}{c}{DVS-CIFAR10} & \multicolumn{2}{c}{DVS128Gesture} & \multicolumn{2}{c}{N-Caltech101} \\
        \cmidrule(lr){3-4} \cmidrule(lr){5-6} \cmidrule(lr){7-8}
         &  & T Step & Acc & T Step & Acc & T Step & Acc \\
        \hline
        LIAF-Net\cite{wu2021liaf} & \XSolidBrush & 10 & 70.4 & 60 & 97.6 & - & - \\
        TA-SNN\cite{yao2021temporal} & \XSolidBrush & 10 & 72.0 & 60 & 98.6 & - & - \\
        ECSNet\cite{chen2022ecsnet} & \XSolidBrush & - & 72.7 & - & 98.6 & - & 69.3 \\
         \hline
        Rollout\cite{kugele2020efficient} & \Checkmark & 48 & 66.8 & 240 & 97.2 & - & - \\
        DECOLLE\cite{kaiser2020synaptic} & \Checkmark & - & - & 500 & 95.5 & - & - \\
        tdBN\cite{zheng2021going} & \Checkmark & 10 & 67.8 & 40 & 96.9 & - & - \\
        PLIF\cite{fang2021incorporating} & \Checkmark & 20 & 74.8 & 20 & 97.6 & - & - \\
        SEW-ResNet\cite{fang2021deep} & \Checkmark & 16 & 74.4 & 16 & 97.9 & - & - \\
        Dspike\cite{li2021differentiable} & \Checkmark & 10 & 75.4 & - & - & - & - \\
        SALT\cite{kim2021optimizing} & \Checkmark & 20 & 67.1 & - & - & - & - \\
        DSR\cite{meng2022training} & \Checkmark & 10 & 77.3 & - & - & - & - \\
        mMND\cite{she2021sequence} & \Checkmark & - & - & - & 98.0 & - & 71.2 \\
        Spikformer\cite{zhou2022spikformer} & \Checkmark & 16 & 80.9 & 16 & 98.3 & - & - \\
        \hline
        \textbf{SEW-ResNet-18(ours)} & \Checkmark & 16 & 81.5 & 16 & 97.9 & 16 & \textbf{84.4} \\
        \textbf{Spikformer-2-256(ours)} & \Checkmark & 16 & \textbf{83.6} & 16 & \textbf{98.62} & 16 & 81.6 \\
        \bottomrule
    \end{tabular}
    \label{compare with SOTA}
\end{table}

The experimental results show that both SEW-ResNet-18 and Spikformer-2-256 exhibit superior performance on DVS-CIFAR10, DVS128Gesture, and N-Caltech101 when employing NeuroMoCo compared to their counterparts without NeuroMoCo.  This indicates that our proposed NeuroMoCo method is effective and structurally compatible for SNNs' pre-training.
\subsubsection{Comparative Experiment}
In this section, we undertake a series of comparative experiments with the aim of validating the performance of our approach and accentuating its advantages.

The performance of provided NeuroMoCo is evaluated on DVS-CIFAR10, DVS128Gesture, and N-Caltech101 using SEW-ResNet-18 and Spikformer-2-256 models. Additionally, a variety of SOTA methods are compared.   The classification performance of our NeuroMoCo and current SOTA methods is presented in Table 4, which shows that our NeuroMoCo obtains remarkable results across all three datasets.

Specifically, on DVS-CIFAR10, our SEW-ResNet-18 and Spikformer-2-256 models achieve classification accuracies of 81.5\% and 83.6\% using 16 time steps, respectively, surpassing the accuracies of 78.0\% and 80.9\% achieved by the same models when randomly initialized (as shown in Table 3). Furthermore, compared to the state-of-the-art Spikformer, our Spikformer-2-256 achieves a 2.7\% improvement in accuracy while employing the same model and time step.  It is worth noting that we outperform loss-based TET (83.2\%), which is not included in the table as it is based on loss rather than network architecture.  This underscores that we have achieved state-of-the-art (SOTA) performance.

For DVS128Gesture, our SEW-ResNet-18 and Spikformer-2-256 models use 16 time steps to achieve a classification accuracy of 97.9\% and 98.62\%, respectively, which is better than 95.8\% and 96.8\% for random initialization of the same model (see Table 3). In addition, compared to the previous state-of-the-art TA-SNN model (60 time steps, 98.6\%), our Spikformer-2-256 uses fewer time steps (16) to achieve higher classification accuracy (98.62\%), which is also the current optimal performance.

Finally, on N-Caltech101, still utilizing 16 time steps, we achieve classification accuracies of 84.4\% and 81.6\% for SEW-ResNet-18 and Spikformer-2-256, respectively.    These accuracies outperform the results of 80.5\% and 79.8\% achieved by the same models with random initialization (as seen in Table 3). Notably, SEW-ResNet-18 (84.4\%) using our NeuroMoCo method exhibits significant improvements of 15.1\% and 13.2\%, respectively, compared to the previously known advanced works ECSNet (69.3\%) and mMND (71.2\%).    To the best of our knowledge, this also represents the current SOTA.

Our models consistently demonstrate performance advantages across all three neuromorphic datasets, underscoring the generality of our proposed NeuroMoCo method for neuromorphic data.  This holds significant implications for numerous practical applications.
\section{Conclusion}
In this paper, a SNN-oriented learning method NeuroMoCo has been introduced to increase the performance on complex neuromorphic datasets, which can be used as a self-supervised pre-training framework for spiking convolution and spiking Transformer structures. 
This is the first instance of applying SSL based on momentum contrastive learning to SNNs. 
In order to further improve the classification accuracy, we have designed a new loss function named MixInfoNCE based on the temporal characteristics of neuromorphic datasets. 
The effectiveness of MixInfoNCE and NeuroMoCo have been validated by extensive ablation experiments. 
After pre-training by NeuroMoCo, Spikformer-2-256 has achieved SOTA performance on DVS-CIFAR10 (83.6\%) and DVS128Gesture (98.62\%), and SEW-ResNet-18 has achieved SOTA performance on N-Caltech101 (84.4\%), which means that SSL is an effective solution  for complex tasks in the field of SNNs to some extent.

\section*{Acknowledgment}
This work was supported by Natural Science Foundation of Chongqing (Grant No. cstc2021jcyj-msxmX0565), Fundamental Research Funds for the Central Universities (Grant No. SWU021002), Project of Science and Technology Research Program of Chongqing Education Commission (Grant No. KJZD-K202100203), and National Natural Science Foundation of China (Grant Nos. U1804158, U20A20227).



\bibliographystyle{elsarticle-harv} 
\bibliography{bib-ref}



\end{document}